\documentclass{article}
\usepackage{spconf,amsmath,graphicx}
\usepackage{booktabs}
\usepackage{graphicx}
\usepackage{tabularx}
\usepackage{amsmath}
\usepackage{amssymb}
\usepackage[utf8]{inputenc}
\usepackage{subfigure}
\usepackage{algorithm}
\usepackage{algpseudocode}
\usepackage{float}
\usepackage{xcolor}
\usepackage{array,multirow,graphicx}
\usepackage{tikz}
\usepackage{amsfonts}
\usepackage{enumitem}
\usepackage[belowskip=-5pt,aboveskip=10pt]{caption}

\let\OLDthebibliography\thebibliography
\renewcommand\thebibliography[1]{
  \OLDthebibliography{#1}
  \setlength{\parskip}{0pt}
  \setlength{\itemsep}{0pt plus 0.3ex}
}

\newcommand{\yf}[1]{\textcolor{black}{{#1}}}

\title{enhancing low-light images using infrared encoded images}
\name{Shulin Tian$^{1*}$, Yufei Wang$^{1*}$, Renjie Wan$^2$, Wenhan Yang$^3$, Alex C. Kot$^1$, Bihan Wen$^{1\dagger}$}
\address{
$^1$ Nanyang Technological University, 
$^{2}$ Hong Kong Baptist University, 
$^{3}$ Peng Cheng Laboratory
\thanks{$*$ means equal contribution, and $\dagger$ means corresponding author}
\thanks{This work was done at Rapid-Rich Object Search (ROSE) Lab, Nanyang Technological University. This research is supported in part by the NTU-PKU Joint Research Institute (a collaboration between the Nanyang Technological University and Peking University
that is sponsored by a donation from the Ng Teng Fong
Charitable Foundation), the Basic and Frontier Research
Project of PCL, the Major Key Project of PCL, and the
MOE AcRF Tier 1 (RG61/22) and Start-Up Grant.}
}

\begin{document}
\maketitle

\begin{abstract}
Low-light image enhancement task is essential yet challenging as it is ill-posed intrinsically.
Previous arts mainly focus on the low-light images captured in the visible spectrum using pixel-wise loss, which limits the capacity of recovering the brightness, contrast, and texture details due to the small number of income photons.
In this work, we propose a novel approach to increase the visibility of images captured under low-light environments by removing the in-camera infrared (IR) cut-off filter, which allows for the capture of more photons and results in improved signal-to-noise ratio due to the inclusion of information from the IR spectrum. To verify the proposed strategy, we collect a paired dataset of low-light images captured without the IR cut-off filter, with corresponding long-exposure reference images with an external filter.
\yf{The experimental results on the proposed dataset demonstrate the effectiveness of the proposed method, showing better performance quantitatively and qualitatively.} The dataset and code are publicly available at \texttt{https://wyf0912.github.io/ELIEI/}

\end{abstract}
\section{Introduction\vspace{-0.5em}}
\label{sec:intro}
\yf{Due to the small number of photons captured by the camera, the images captured under low-light environments usually suffer from poor visibility, intense noise, and artifacts. To enhance the visibility of the images captured in low-light environments, previous works mainly focus on modelling the mapping 
relationship between low-light images and corresponding normally-exposed images.}
%
%
%
%
\yf{Specifically, current deep learning based} methods have the following paradigms: learning an end-to-end model using paired datasets in~\cite{chang2015retinex, shen2017msr, wei2018deep, park2018dual, wang2023removing}; GAN-based networks in~\cite{chen2018deep, jiang2021enlightengan}; encoder-decoder based models in~\cite{lore2017llnet, chen2018encoder, ren2019low, wang2023raw}. However, the aforementioned methods are all based on existing visible information of the corrupted inputs on RGB space,
\yf{\textit{i.e.}, even if they can achieve pleasant perceptual quality, they can not perform reliably due to the lack of incident photons~\cite{goyal2021photon}.} 
Besides, there are various limitations of the current mainstream methods, \textit{e.g.}, 
\yf{end-to-end training using pixel reconstruction loss leads to a regression-to-mean problem}; GAN-based training requires careful hyper-parameter tuning \yf{and lacks enough supervision for noise removal. }


Recently, infrared-light-based methods have attracted great attention in low-level computer vision tasks
\yf{as they introduce extra information from infrared spectroscopy.}
There are several works explored the usage of infrared light in computation photography previously. \yf{Specifically, Zhuo \textit{et al.}~\cite{NIRflash} propose} to use additional Near-Infrared (NIR) flash images instead of normal flash images to restore the details of noisy input images that require the user to take two photos of the same scene in a static environment, causing the misalignment of the inputs easily; 
Zhang \textit{et al.}~\cite{zhang2008enhancing} propose a dual-camera system to capture a NIR image and a normal visible image of the same scene concurrently, while increasing the cost of devices during the acquisition of data.

\begin{figure}[tbp]
    \centering
    \includegraphics[width=\linewidth]{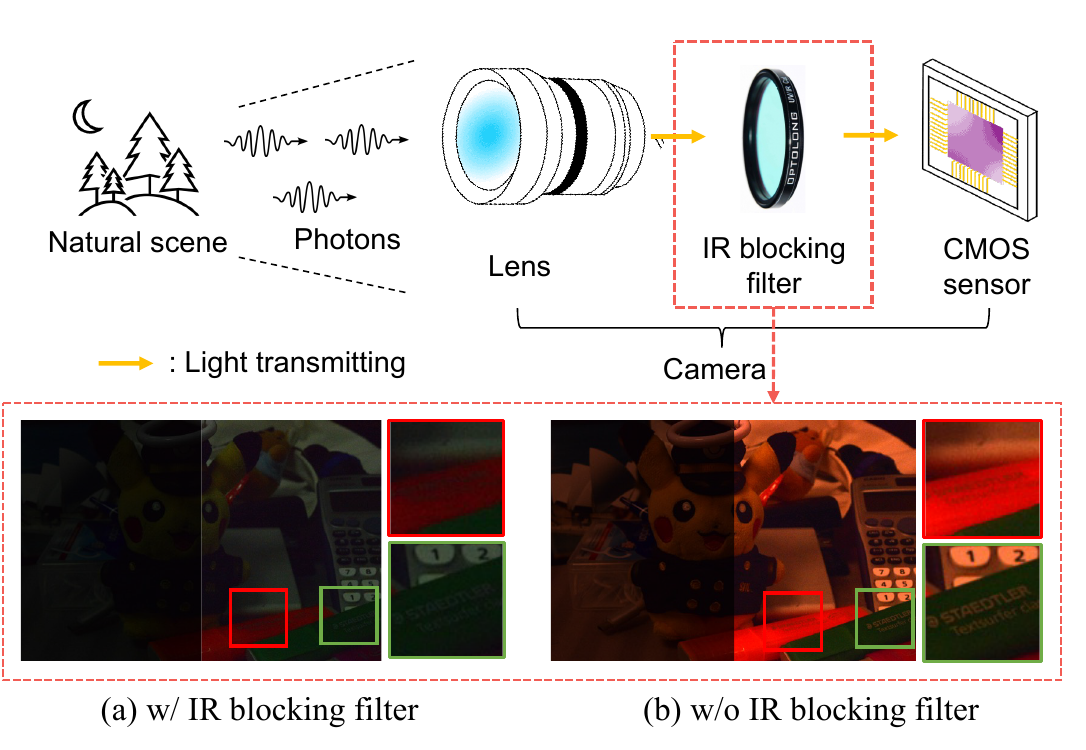}
    \caption{
    The visibility of low-light images is enhanced by increasing the number of income photons. (a) and (b) are captured under the same settings, except whether the IR cut-off filter exists. (The right sides of (a) and (b) are amplified by a factor of 3.5 for better visualization.)}
    \label{fig:intro}
    \vspace{-1em}
\end{figure}

In this paper, we propose a novel prototype that utilizes information from the infrared spectrum without the need for additional devices. Most solid-state (CCD/CMOS) based digital cameras are equipped with IR cutoff filters to avoid color distortion caused by the high sensitivity to IR light. Conversely, we remove the IR cutoff filter so that the CMOS can receive more incident photons located on the infrared spectrum, resulting in increased brightness, higher signal-noise ratio, and improved details as shown in Fig. \ref{fig:intro}. A paired dataset, namely IR-dataset, of IR-RGB images captured under low-light environments and their reference normally-exposed RGB images, is collected under different scenes. We further propose a novel flow-based model that can enhance visibility by modelling the distribution of normally-exposed images and address color distortion caused by the lack of IR cutoff filter through our proposed color alignment loss (CAL).


\yf{In summary,} the contributions of our work \yf{are threefold}:
\begin{enumerate}[itemsep=2pt,topsep=2pt,parsep=0pt]
    \item
    \yf{We collect a paired dataset under a novel prototype, \textit{i.e.,} IR-RGB images captured under low-light environments and their normally-exposed reference RGB images, which supports future studies.}
    \item
    \yf{We propose a flow-based model with our proposed color alignment loss, which can effectively address the color distortion caused by removing the IR-cut filter.}
    \item
    \yf{We conduct extensive experiments on our collected datasets that demonstrate removing the IR-cut filter can lead to better-quality restored images in low-light environments. Besides, our proposed framework achieves superior performance compared with SOTA methods.}
\end{enumerate}

\begin{figure*}[htbp]
    \centering
    \subfigure[Input]{
        \includegraphics[width=0.194\linewidth, height=0.12933\linewidth]{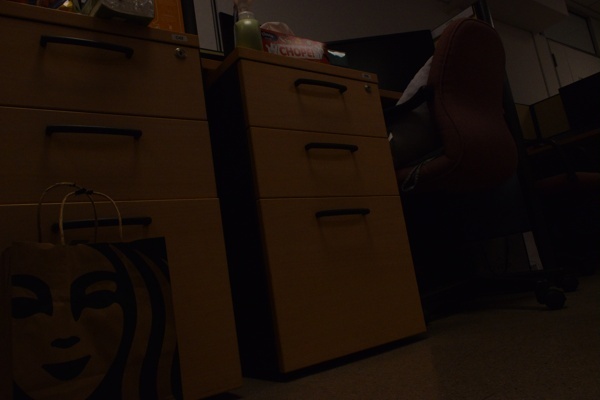}
    }\hspace{-8pt}
    \subfigure[RetinexNet]{ 
        \includegraphics[width=0.194\linewidth, ]{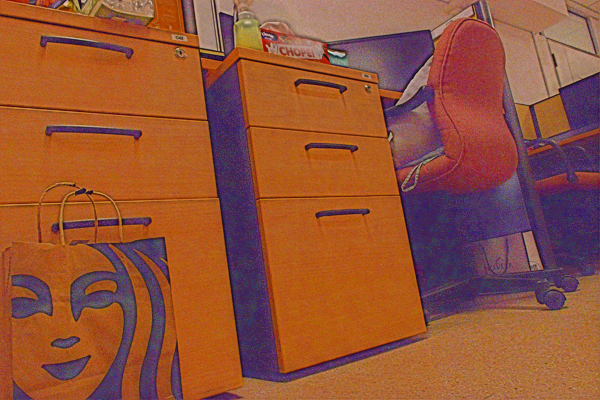}
    }\hspace{-8pt}
    \subfigure[LIME]{ 
        \includegraphics[width=0.194\linewidth, height=0.12933\linewidth]{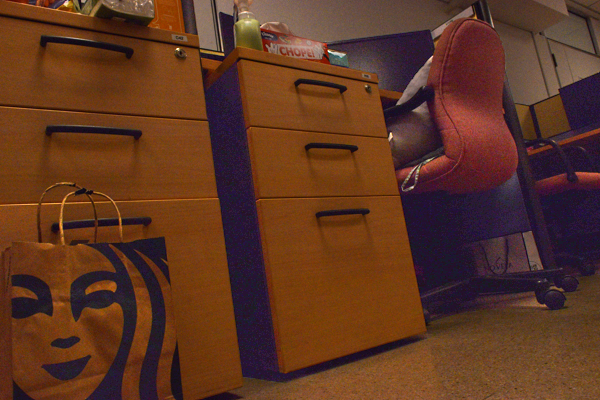}
    
    }\hspace{-8pt}
    \subfigure[Zero-DCE]{ 
        \includegraphics[width=0.194\linewidth, height=0.12933\linewidth]{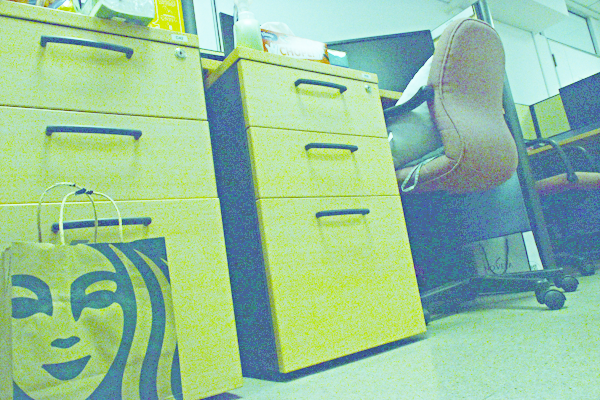}
    }\hspace{-8pt}
    \subfigure[KinD]{ 
        \includegraphics[width=0.194\linewidth, height=0.12933\linewidth]{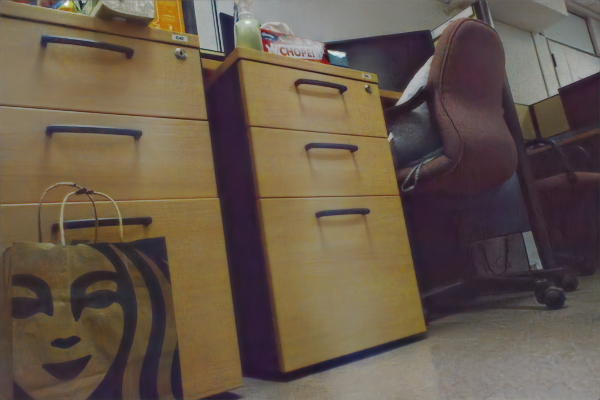}
    \vspace{-2em}
    }
    \\
    \subfigure[KinD++]{ 
        \includegraphics[width=0.194\linewidth, height=0.12933\linewidth]{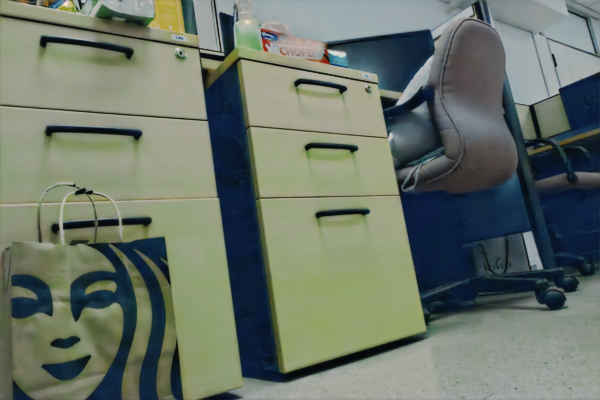}
    }\hspace{-8pt}
    \subfigure[EnlightenGAN]{ 
        \includegraphics[width=0.194\linewidth, height=0.12933\linewidth]{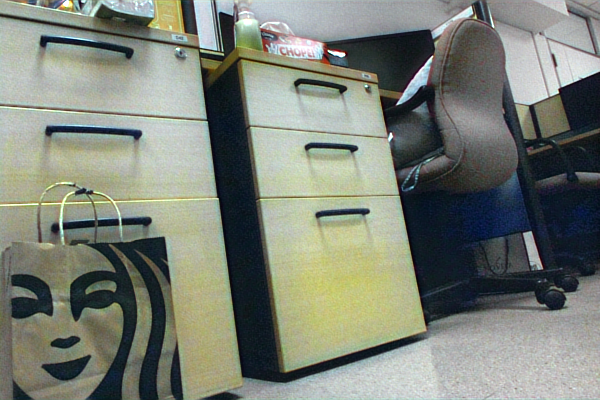}
    }\hspace{-8pt}
    \subfigure[MIRNet]{ 
        \includegraphics[width=0.194\linewidth, height=0.12933\linewidth]{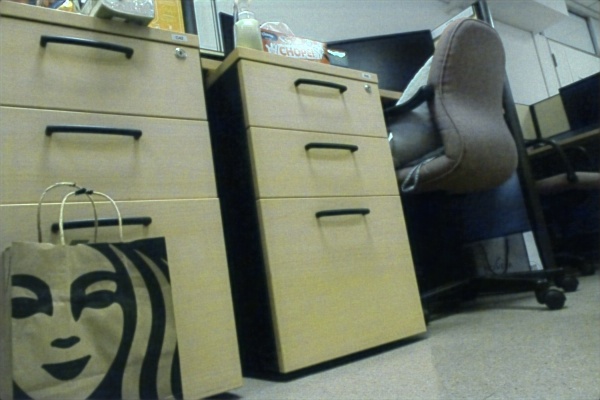}
    }\hspace{-8pt}
    \subfigure[Ours]{
        \includegraphics[width=0.194\linewidth, height=0.12933\linewidth]{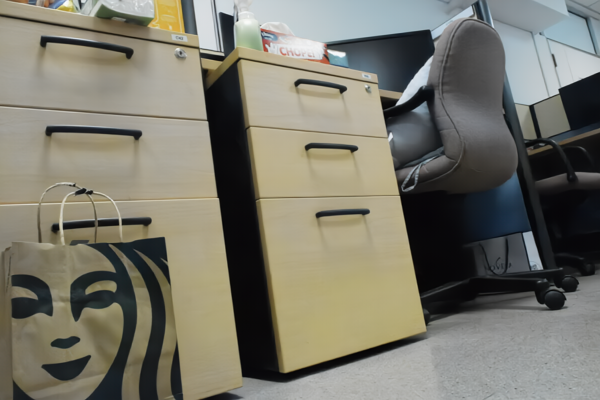}
    }\hspace{-8pt}
    \subfigure[Reference]{
        \includegraphics[width=0.194\linewidth, height=0.12933\linewidth]{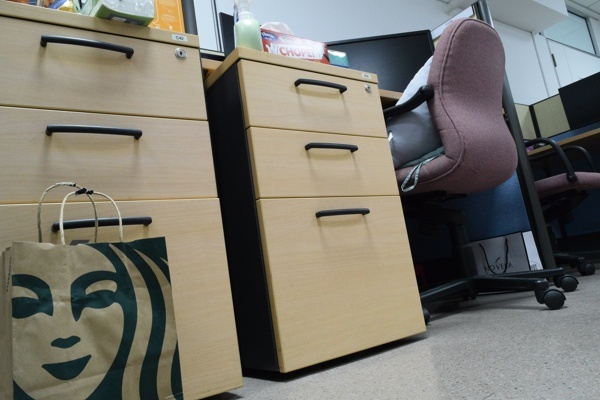}
    }
    \vspace*{-1.5mm}
    \caption{Visual comparison with state-of-the-art low-light image enhancement methods on IR-RGB dataset. Our method shows better performance in controlling color distortion and \yf{detail preservation}.}
    \label{fig:indoor_res}
    \vspace{-0.4cm}
\end{figure*}


\vspace{-1em}
\section{Methodology\vspace{-0.5em}}
\label{sec:enhancement}

\subsection{Dataset Collection\vspace{-0.5em}}
\label{ssec:dataset}

 
The dataset is collected by a modified Nikon D3300 camera, in which the internal IR cut-off filter is removed.
The paired images are captured using a stable tripod and remote control to minimize misalignment. The low-light images are captured using the aforementioned device without IR cut-off filter. To capture the normally-exposed reference images in the visible light spectrum, an external IR filter, which has the same cut-off wavelength as the internal one, is carefully put in front of the lens to ensure that no camera shift occurs during the long exposure. To better explore the effectiveness of removing the IR cut-off filter in a low-light environment, we also collect a set of low-light images in the visible light spectrum (\textit{e.g.}, the example in Fig.~\ref{fig:intro}).
We divide our dataset into a training set and an evaluation set. Specifically, the training set includes 236 pairs of low-light images without cut-off filter and their corresponding reference images (472 images in total). The evaluation set has 80 pairs of low-light images with and without the cut-off filter and their corresponding reference images.

\vspace{-1em}
\subsection{Preliminary\vspace{-0.5em}}
\label{ssec:pre}
Previously, the mainstream of deep learning based models is mainly based on pixel reconstruction loss. However, due to the limited capacity to distinguish the unwanted artifacts with the real distribution of normally-exposed images, they may lead to unpleasant visual quality with blurry outputs~\cite{zhang2021beyond, wang2021low}.



Inspired by the extraordinary performance of flow-based models~\cite{lugmayr2020srflow, xiao2020invertible, wolf2021deflow, wang2021low}, we found that learning conditional probability distribution can handle the aforementioned problem by including possibilities of various distributions of natural images. Specifically, the recent state-of-the-art LLFlow model~\cite{wang2021low} has shown great performance in using normalizing flow conditioned on corrupted inputs to capture the conditional distribution of normally exposed images. In this work, we inherited the 
\yf{core idea} of conditional flow with the likelihood estimation proposed in~\cite{wang2021low} \yf{as the backbone of our method.}
\yf{The conditional probability density function of normally exposed images can be modified as follows:}
\begin{equation}
\label{eqn:condPDF}
\setlength{\abovedisplayskip}{3pt}
\setlength{\belowdisplayskip}{3pt}
    f_{cond}(y|x) = f_z(\Theta(y;x))|\det\frac{\partial\Theta}{\partial y}(y;x)|,
\end{equation}
where $\Theta(\cdot)$ is the invertible network with $N$ invertible layers $\{\theta ^1, \theta ^2, \dots, \theta ^N\}$, and the latent representation $z=\Theta(y;x)$ is mapped from the corrupted inputs $x$ normally exposed images $y$. By characterizing the model with maximum likelihood estimations, the model can be optimized with the negative log-likelihood loss function:

\begin{equation}
\setlength{\abovedisplayskip}{-3pt}
\setlength{\belowdisplayskip}{0pt}
\begin{split}
\small
    \mathcal{L}_{nll}&(x, y) = -\log f_{cond}(y|x) \\
        &=  -\log f_z(\Theta (y; x)) 
        - \sum_{n=0}^{N-1} \log |\det 
         \frac{\partial \theta^n}{\partial z^n}(z^n; g^n(x_l))|,
\end{split}
\end{equation}
where $g(\cdot)$ is the encoder that outputs conditional features of the layers $\theta ^i$ from the invertible network.

\vspace{-1em}
\subsection{Color Alignment Loss\vspace{-0.5em}}
\label{ssec:caloss}
Although the benchmarks performed well on the visible light spectrum, the performance suffered from severe degradation caused by the additional infrared light in some extreme cases if we directly apply benchmark methods to the collected dataset. To further alleviate the color distortion caused by removing the IR filter, inspired by histogram-matching techniques studies~\cite{neumann2005color, rakwatin2007stripe, rakwatin2008restoration}, used by remote sensing, we propose to minimize the divergence of the color distribution between the generated images and reference images. Specifically, by representing the color information using differentiable histograms in the RGB color channels, we emphasize more on the color distributions of the generated and reference images instead of the local details. To further measure the differences in these distributions, we propose using the Wasserstein distance, which can provide a more stable gradient compared with the commonly used KL divergence. The details are as follows:
\vspace{-10pt}
\subsubsection{Differentiable Histogram\vspace{-0.5em}}
Since the low-light images are taken without the existence of an IR cut-off filter, they admit more red light, which leads to color bias in the red channel. To suppress the color distortion, we propose to minimize the divergence of the channel-wise differentiable histogram between the generated and reference images.
Assume that $x\in \mathbb{R}^{C \times H \times W}$ is an image where $C$, $H$ and $W$ refer to its number of channels, height, and width respectively.

To calculate \yf{its} channel-wise histogram bounded by an arbitrary range $[a;b]$, we consider fitting the histogram with uniformly spaced bins with size $R$, noted by nodes $t_i \in \{t_1 = m, t_2, \dots, t_R = n\}$, where step size $\Delta = \frac{(a-b)}{R-1}$. By matching the pixel values of different channels of the image to the histogram nodes, the value $h_r$ of the histogram $H$ at each node then be calculated as:
\begin{equation}
\label{eqn:hist}
\setlength{\abovedisplayskip}{3pt}
\setlength{\belowdisplayskip}{3pt}
    h_r = \sum_{C} \frac{1}{1+\delta*(p_{i,j}-t_r)^2}, r = 1,2,\dots, R
\end{equation}
where $\delta$ is a constant scaling factor. \yf{After} collating and normalizing $h_r$, we could get the final one-dimensional histogram $H(x)$ with size $R$ on different channels.
\vspace{-1em}
\subsubsection{Wasserstein Metric\vspace{-0.5em}}
\label{sssec: emd}

Inspired by Wasserstein distance (W-distance) to measure the distance between distributions on a given metric space~\cite{chen2022computing}, we propose to optimize the histograms of images using W-distance as follows
\begin{equation}
\label{eqn:wdis_our}
\setlength{\abovedisplayskip}{3pt}
\setlength{\belowdisplayskip}{3.5pt}
W_p (H_{\hat{y}}, H_y) = \inf_{\hat{y}\sim H_{\hat{y}}, y\sim H_y}(\mathbb{E}||\hat{y}-y||^p)^{1/p},
\end{equation}
\yf{where $H_{\hat{y}}$ and $H_y$ denote differentiable histograms of the restored image $\hat{y}$ and ground-truth image $y$ respectively through Eq. (\ref{eqn:hist}).}
An explicit formula can be obtained since the dimension of the variable is $1$ as follows,
\begin{equation*}
\setlength{\abovedisplayskip}{3pt}
\setlength{\belowdisplayskip}{3.5pt}
\begin{split}
W_p (H_{\hat{y}}, H_y) &= ||\yf{F}_{\hat{y}}^{-1} - \yf{F}_y^{-1}||_p  \\
&= (\int_{a}^{b} |\yf{F}_{\hat{y}}^{-1}(\alpha) - \yf{F}_y^{-1}(\alpha)|^p \mathrm{d} \alpha)^{1/p},
\end{split}
\end{equation*}
\yf{where $F_{y}$ and $F_{\hat{y}}$ are the cumulative distribution of $H_{y}$ and $H_{\hat{y}}$ respectively.} It could be further simplified when $p=1$ \yf{and the variable is discrete}:
\begin{equation}
\setlength{\abovedisplayskip}{3pt}
\setlength{\belowdisplayskip}{3.5pt}
\mathcal{L}_{CA} = W_1(H_{\hat{y}}, H_y)
= \sum_{\mathbb{R}} |\yf{F}_{\hat{y}}(t)-\yf{F}_y(t)|\mathrm{d} t.
\end{equation}
The negative log-likelihood and the color alignment loss jointly define the total loss as follows
\begin{equation}
\label{eqn:totaloss}
\setlength{\abovedisplayskip}{3pt}
\setlength{\belowdisplayskip}{3pt}
    \mathcal{L} = \mathcal{L}_{nll} + \lambda \cdot \mathcal{L}_{CA},
\end{equation}
where $\lambda$ is a weighting constant to adjust the scales of color alignment loss for specific settings.
\vspace{-0.5em}
\section{Experiments\vspace{-0.5em}}
\label{sec:experiments}
\vspace{-0.5em}
\subsection{Experimental settings.\vspace{-0.5em}}
All the captured images are resized to the resolution of $400\times600$ \yf{for training and testing}. 
For our model, the \yf{weighting factor $\lambda$} of CAL \yf{is set to} $0.01$ to cast the loss component value onto \yf{a similar} numerical scale during training;
to simplify the task, we bound \yf{the range of} the channel-wise histogram values \yf{to} $[0.0;1.0]$, and the bin size is \yf{set} to $64$ per channel. 

\vspace{-1em}
\subsection{Evaluations results.\vspace{-0.5em}}
\yf{To evaluate the performance of different methods on the proposed dataset, we retrain all the methods using the same training data, \textit{i.e.,} the training set of our proposed dataset. For a fair comparison, we explore training hyper-parameters of competitors in a wide range and report the best performance we obtained. We report the experimental results in Table \ref{tab:comp} and visual comparison in Fig. \ref{fig:indoor_res}.}
Based on our evaluation and analysis of the experiment results
\yf{As we can see in the table}, Retinex-theory-based methods \yf{exhibit} limited generalization ability and unpleasant outputs, \textit{e.g.}, RetinexNet~\cite{wei2018deep}, Kind~\cite{zhang2019kindling}, KinD++~\cite{zhang2021beyond}. \yf{We conjecture the reason is that} the aforementioned methods assume the existence of an invariant reflectance map across low-light inputs and ground truth images and require a shared network to extract both illumination and reflectance maps \yf{of them}
, which is not feasible \yf{in our setting}. \yf{Besides, our method achieves the best performance among all competitors in terms of both fidelity and perceptual quality. }

\begin{table}[tb]
    \centering
    \scalebox{1}{
    \begin{tabular}{cccccc}
    \toprule
        \ & PSNR $\uparrow$ &  SSIM $\uparrow$ & LPIPS $\downarrow$\\
    \midrule
        RetinexNet~\cite{wei2018deep} & 11.14 & 0.628 & 0.586 \\
        LIME~\cite{guo_lime} & 11.31 & 0.639 & 0.560\\
        Zero-DCE~\cite{guo2020zero} & 11.40 & 0.592 & 0.443 \\
        KinD~\cite{zhang2019kindling} & 14.73 &  0.714 & 0.357\\
        EnlightenGAN~\cite{jiang2021enlightengan} & 16.95 &  0.715 & 0.357\\
        KinD++~\cite{zhang2021beyond} & 17.84 &  0.830 & 0.249\\
        MIRNet~\cite{Zamir2020MIRNet} & 22.23 &  0.833 & 0.224\\
        LLFlow~\cite{wang2021low} & 25.46 & 0.890 & 0.130 \\
        Ours & \textbf{26.23} & \textbf{0.899} & \textbf{0.116}\\
    \bottomrule
    \end{tabular}
    }
    \vspace{-1.5mm}
    \caption{Quantitative comparison of existing SOTA methods \yf{and} our method on the proposed dataset. \yf{We adapt PSNR, SSIM, and LPIPS to measure the reconstruction quality, structural similarity, and perceptual quality respectively}; $\uparrow(\downarrow)$ means higher(lower) values stand for better quality. }
    \label{tab:comp}
\end{table}

\vspace{-1em}
\subsection{Ablation Study\vspace{-0.5em}}
\label{ssec:ablation}




\begin{figure}[tbp]
\begin{minipage}[b]{\linewidth}
  \centering
  \centerline{\includegraphics[width=\linewidth]{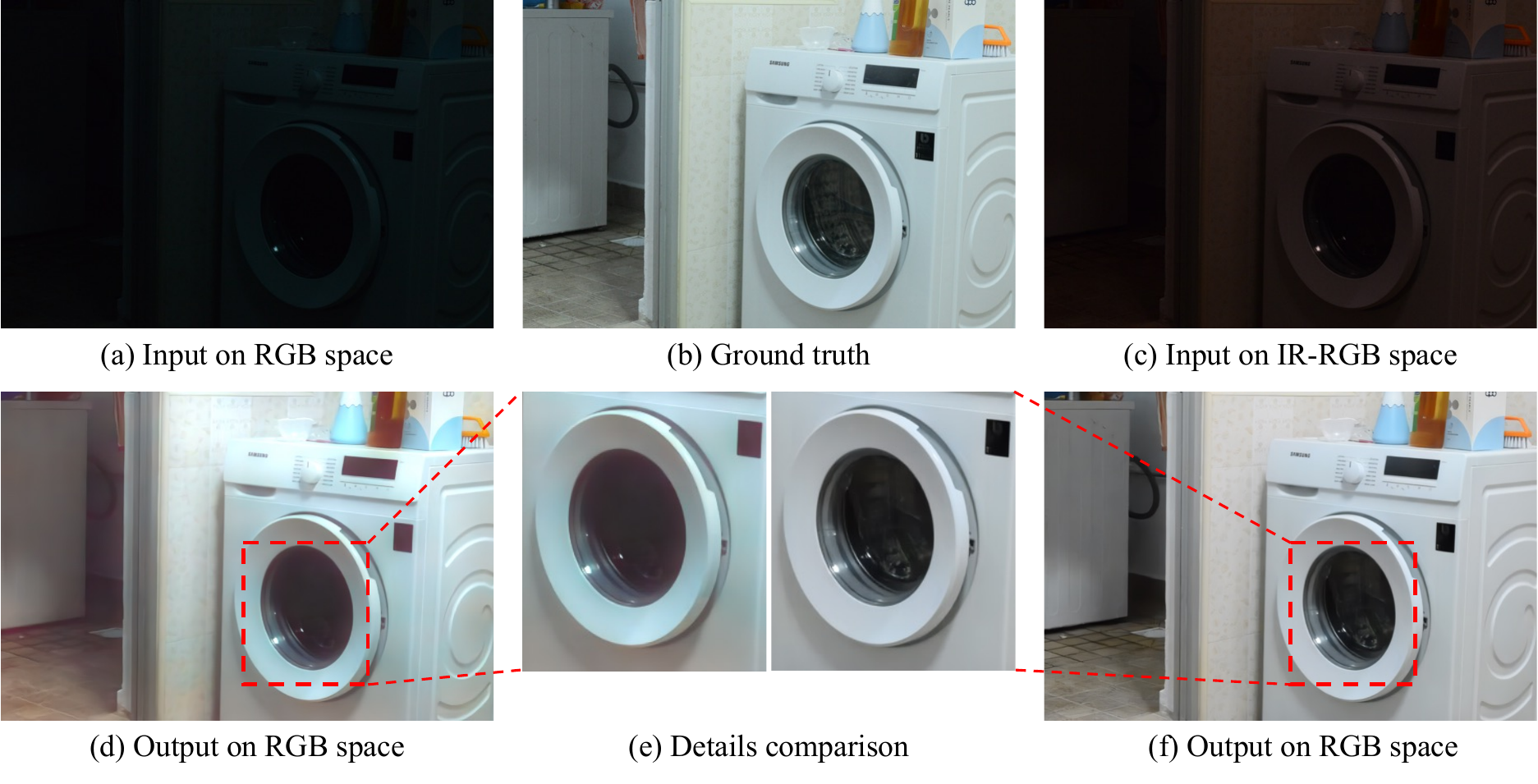}}
\vspace{-0.5cm}
\end{minipage}
\vspace*{-1.5mm}
\caption{Visual comparisons of pretrained benchmark RGB-model with our IR-model. (a), (c) are images captured from RGB and IR-RGB space separately under low-light conditions, (d), (f) are respective high-light outputs.}
\label{fig:abl1}
\end{figure}

\textbf{1) \yf{Effectiveness of removing IR cut-off filter}.} To \yf{further} verify the \yf{effect} of \yf{removing the internal IR cut-off filter}, we compare both quantitative and visual results that were restored from standard RGB space and IR-RGB space separately. \yf{For the models evaluated on the visible light spectrum, we utilize the pretrained/released models from SOTA methods trained on a large-scale dataset so that they have good generalization ability to different scenarios.}
As shown in Table \ref{tab:comp_abl}, the quantitative results calculated from IR light encoded image with our model are much higher than those directly restored from standard \yf{visible light spectrum}. \yf{Besides, for the same method, especially for the method utilizing fully supervised training manner, there exists an obvious performance gap by converting the input space from IR-visible spectrum to only visible spectrum, which demonstrates that removing the IR cut-off filter may lead to the higher noise-signal ratio in extreme dark environment.}
Besides, as shown in Fig. \ref{fig:abl1}, the \yf{reconstructed image} with IR light performs better in recovering local features and details of the image.

\begin{figure}[tb]

\begin{minipage}[b]{.48\linewidth}
  \centering
  \centerline{\includegraphics[width=\linewidth]{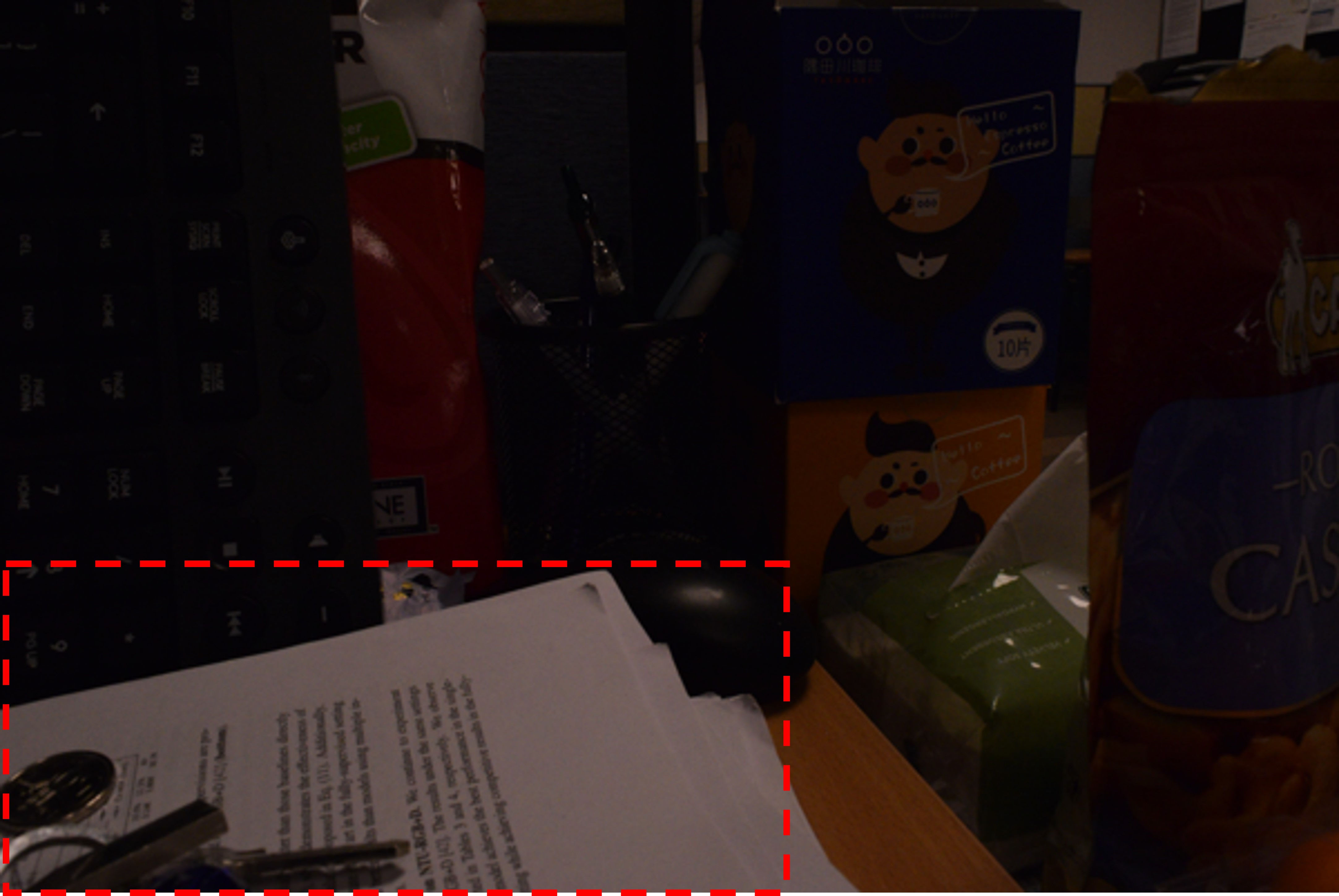}}
  \centerline{(a) Input}\medskip
\end{minipage}
\hfill
\begin{minipage}[b]{.48\linewidth}
  \centering
  \centerline{\includegraphics[width=\linewidth]{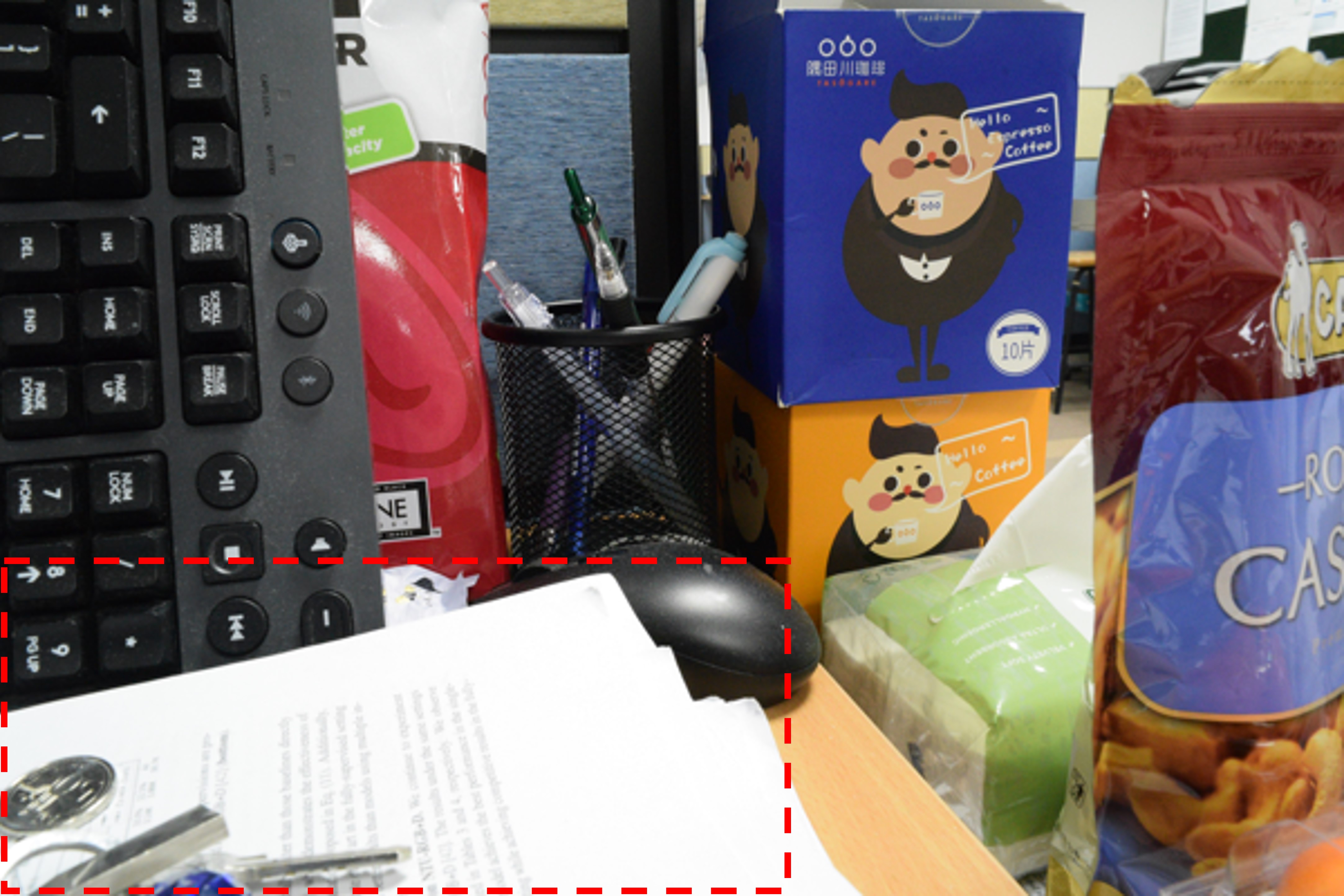}}
  \centerline{(b) Reference}\medskip
\end{minipage}
\begin{minipage}[b]{0.48\linewidth}
  \centering
  \centerline{\includegraphics[width=\linewidth]{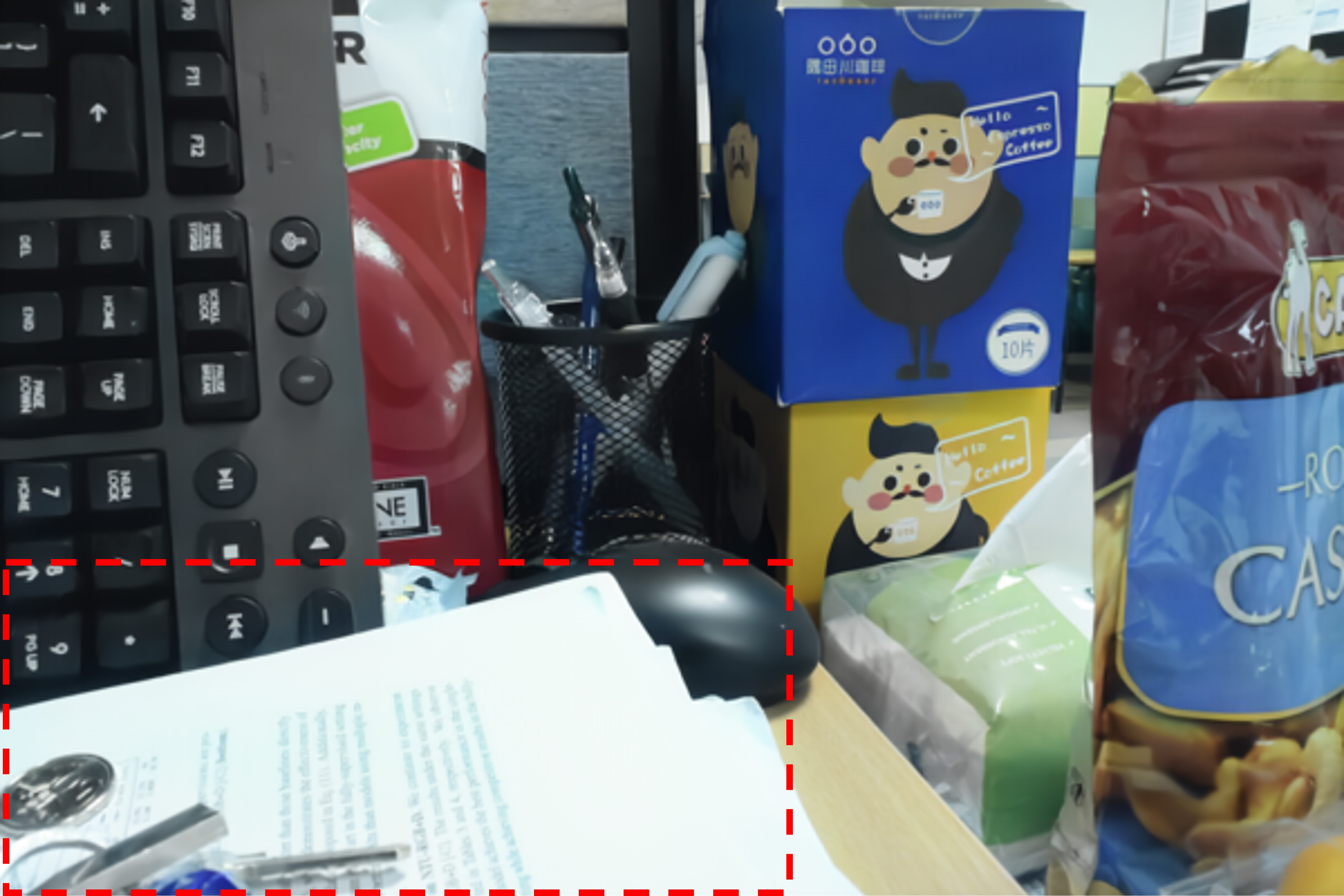}}
  \centerline{(c) w/o CAL}\medskip
\end{minipage}
\hfill
\begin{minipage}[b]{.48\linewidth}
  \centering
  \centerline{\includegraphics[width=\linewidth]{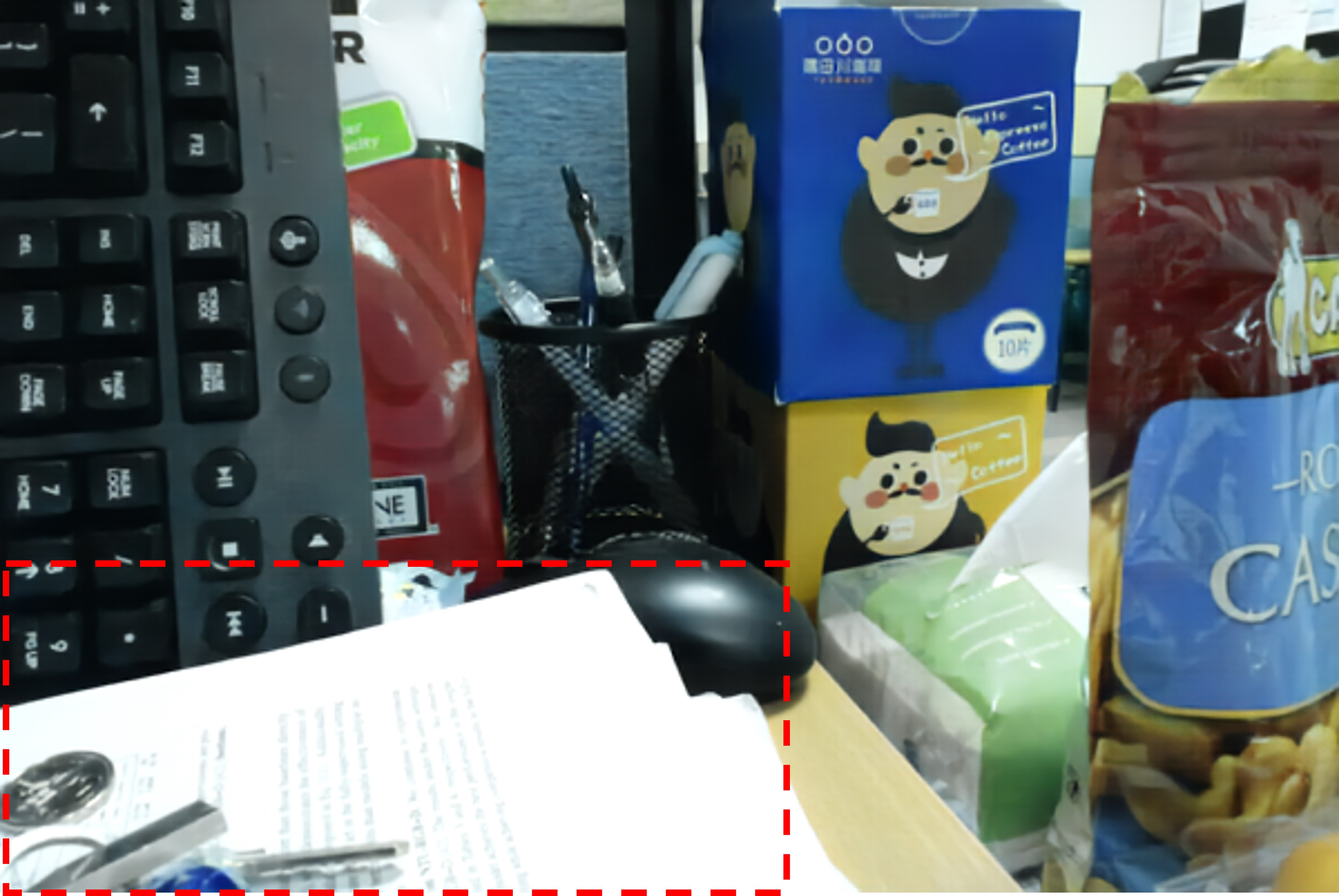}}
  \centerline{(d) w/ CAL}\medskip
\end{minipage}
\vspace*{-5mm}
\caption{Comparison of the effectiveness of using CAL. (c) is the result of our model trained directly w/o adding CAL, and (d) is the output from the same architecture but w/ CAL. }
\label{fig:abl2}
\end{figure}

\begin{table}[t]
    \centering
    \scalebox{1}{
    \begin{tabular}{ccccc}
    \toprule
        \ & PSNR $\uparrow$ &  SSIM $\uparrow$ & LPIPS $\downarrow$\\
    \midrule
        LIME~\cite{guo_lime} & 12.17 & 0.585 & 0.552\\
        Zero-DCE~\cite{guo2020zero} & 12.62 & 0.637 & 0.474\\
        EnlightenGAN~\cite{jiang2021enlightengan} & 13.07 & 0.603  & 0.566\\
        KinD~\cite{zhang2019kindling} & 14.01 & 0.668  & 0.421\\
        RetinexNet~\cite{wei2018deep} & 14.05 & 0.554 & 0.640\\

        KinD++~\cite{zhang2021beyond} & 14.35 & 0.701 & 0.366\\
        
        MIRNet~\cite{Zamir2020MIRNet} & 16.46 & 0.737 & 0.370\\
        LLFlow~\cite{wang2021low} & 19.02 & 0.778 & 0.354\\
        Ours & \textbf{26.23} & \textbf{0.899} & \textbf{0.116}\\

    \bottomrule
    \end{tabular}
    }
    \caption{Quantitative comparison of pretrained SOTA methods on a large-scale visible light spectrum with our method trained on the proposed dataset. $\uparrow(\downarrow)$ means higher(lower) values stand for better quality.}
    \label{tab:comp_abl}
\end{table}
\noindent \textbf{2) The \yf{effectiveness} of color alignment loss.} To validate the assumption of using color alignment loss can improve the imaging quality, we compare the visual quality difference of the usage of color alignment loss. As shown in Fig. \ref{fig:abl2}, the result with CAL shows better perceptual quality with aligned color correctness and higher contrast. However, the original method without CAL appears to have obvious color distortion and blurry edges.

\vspace{-1.5em}
\section{Conclusion\vspace{-0.5em}}
\label{sec:conclusion}
In this paper, we present a novel strategy for tackling low-light image enhancement tasks which introduces more income photons in the IR spectrum. The proposed prototype leads to a higher noise signal ratio in the extreme-dark environment. Based on the proposed prototype, a paired dataset is collected under different scenarios.
%
Experimental results on the proposed dataset show our method achieves the best performance in both quantitative results and perceptual quality. Our prototype shed light on the potential new designs for the digital cameras by exploiting the spectroscopic information captured from infrared light spectrum, providing better image quality with more practical solutions for customers.

\vfill\pagebreak
\bibliographystyle{IEEEbib}
\bibliography{strings,refs}

\end{document}